\documentclass[letterpaper]{article} 
\usepackage{aaai25}  
\usepackage{times}  
\usepackage{helvet}  
\usepackage{courier}  
\usepackage[hyphens]{url}  
\usepackage{graphicx} 
\usepackage{natbib}  
\usepackage{caption} 
\usepackage{algorithm}
\usepackage{algorithmic}
\usepackage{makecell}
\usepackage{multirow}
\usepackage{tikz}
\usepackage{tcolorbox}
\usepackage{pifont}
\usepackage{subcaption}
\usepackage{amsmath,amsfonts}
\usepackage{newfloat}
\usepackage{listings}
\DeclareCaptionStyle{ruled}{labelfont=normalfont,labelsep=colon,strut=off} 
\floatstyle{ruled}
\newfloat{listing}{tb}{lst}{}
\floatname{listing}{Listing}
\title{HC-LLM: Historical-Constrained Large Language Models for Radiology Report Generation}
\author{
    Tengfei Liu\textsuperscript{\rm 1},
    Jiapu Wang\textsuperscript{\rm 1},
    Yongli Hu\textsuperscript{\rm 1}\thanks{Corresponding author},
    Mingjie Li\textsuperscript{\rm 2},
    Junfei Yi\textsuperscript{\rm 3},\\
    Xiaojun Chang\textsuperscript{\rm 4},
    Junbin Gao\textsuperscript{\rm 5},
    Baocai Yin\textsuperscript{\rm 1}
}
\affiliations{
    \textsuperscript{\rm 1}School of Information Science and Technology,
    Beijing University of Technology, Beijing, China\\
     \textsuperscript{\rm 2}Stanford University, Palo Alto CA 94305 USA\\
     \textsuperscript{\rm 3}School of Electrical and Information Engineering, Hunan University, Hunan, China\\
     \textsuperscript{\rm 4}School of Information Science and Technology,
     University of Science and Technology of China, Hefei, China\\
     \textsuperscript{\rm 5}University of Sydney Business School, The University of Sydney, Camperdown, NSW 2006, Australia\\
     \{tfliu, jpwang\}@emails.bjut.edu.cn, \{huyongli,ybc\}@bjut.edu.cn, lmj695@stanford.edu, yijunfei@hnu.edu.cn, cxj273@gmail.com, junbin.gao@sydney.edu.au
}

\usepackage{bibentry}

\begin{document}

\maketitle

\begin{abstract}
Radiology report generation (RRG) models typically focus on individual exams, often overlooking the integration of historical visual or textual data, which is crucial for patient follow-ups. Traditional methods usually struggle with long sequence dependencies when incorporating historical information, but large language models (LLMs) excel at in-context learning, making them well-suited for analyzing longitudinal medical data. In light of this, we propose a novel Historical-Constrained Large Language Models (HC-LLM) framework for RRG,  empowering LLMs with longitudinal report generation capabilities by constraining the consistency and differences between longitudinal images and their corresponding reports. Specifically, our approach extracts both time-shared and time-specific features from longitudinal chest X-rays and diagnostic reports to capture disease progression. Then, we ensure consistent representation by applying intra-modality similarity constraints and aligning various features across modalities with multimodal contrastive and structural constraints. These combined constraints effectively guide the LLMs in generating diagnostic reports that accurately reflect the progression of the disease, achieving state-of-the-art results on the Longitudinal-MIMIC dataset. Notably, our approach performs well even without historical data during testing and can be easily adapted to other multimodal large models, enhancing its versatility. Code is available at: \url{https://github.com/TengfeiLiu966/HC-LLM}.
\end{abstract}

\section{Introduction}

\begin{figure}[t]
\centering
\includegraphics[width=0.95\linewidth]{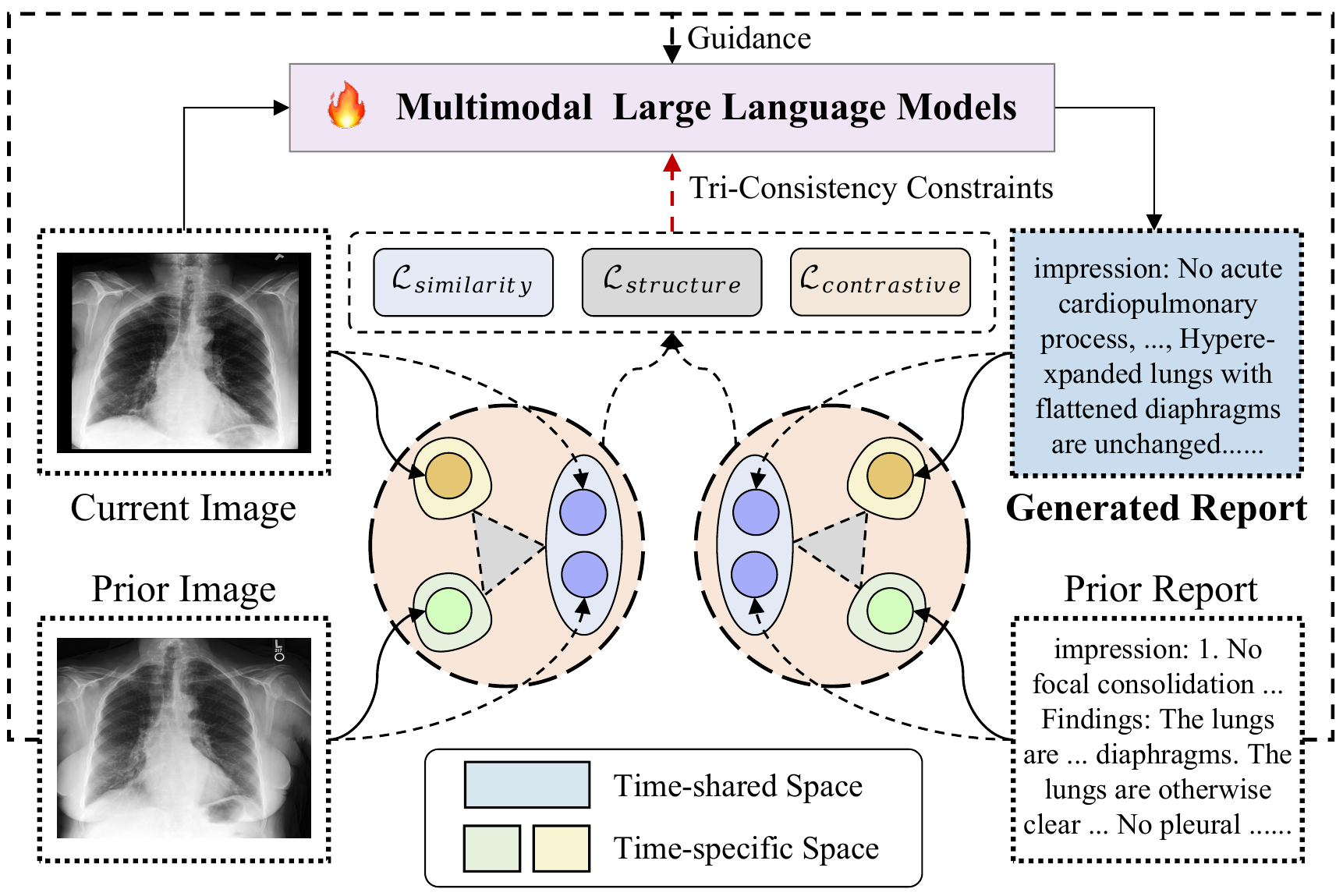}
\vspace{-2mm}
\caption{Illustration of the longitudinal report generation process. Unlike traditional methods that rely solely on the current chest X-ray, our approach emphasizes the effective utilization of historical diagnostic information to significantly enhance the accuracy of LLMs in RRG.}
\label{fig 1}
\vspace{-5mm}
\end{figure}

Radiology report generation (RRG) is a crucial research area in medical AI, with numerous studies focused on reducing the heavy workload of radiologists. In clinical practice, an essential function of these reports is to document pathological changes in patients, aiding doctors in recalling and diagnosing disease progression. As a result, ground truth reports often include descriptions of historical information. However, most existing models \cite{chen2020generating, liu2021exploring, chen2022cross, li2023dynamic, li2023unify, tanida2023interactive, liu2023systematic, huang2023kiut, wang2023r2gengpt, wang2023metransformer, liu2024context, jin2024promptmrg, liu2024bootstrapping, shen2024automatic,li2024contrastive} rely on a single image as input, preventing them from accurately generating descriptions of prior references and thereby impacting their performance. This limitation is evident as current models struggle to achieve high scores on natural language generation (NLG) metrics. Therefore, as illustrated in Figure~\ref{fig 1}, this paper focuses on a more practical research problem: how to generate radiology reports from longitudinal data.

\citet{zhu2023utilizing} have made preliminary attempts in this direction by utilizing a memory mechanism to incorporate historical information for enhanced chest X-ray report generation. However, their approach still relies on traditional cross-attention mechanisms and requires the presence of historical data during testing, limiting its practicality. Recently, large language models (LLMs) \cite{wang2024large} have been successfully applied to traditional radiology report generation tasks \cite{jin2024promptmrg, liu2024bootstrapping, liu2024context}, and their inherent in-context learning abilities make them well-suited for analyzing longitudinal medical data. However, despite this potential, directly inputting longitudinal medical data into LLMs often struggles to produce reports that accurately capture the progression of diseases over time. Therefore, our work focuses on a key challenge: How can historical diagnostic information be effectively utilized to enhance the radiology report generation capabilities of LLMs?

To address this challenge, we propose a Historical-Constrained Large Language Models (HC-LLM) framework for RRG. Specifically, considering changes in disease progression, which may involve the disappearance, stability, and emergence of conditions, we first extract time-shared (stability) and time-specific (disappearance, emergence) features from chest X-rays and diagnostic reports at two-time points. We then employ two types of constraints: intra-modality and inter-modality constraints, which indirectly guide LLM generation by aligning the shared and specific features of the generated and historical reports with those of the corresponding longitudinal images. This alignment ensures that the generated reports accurately capture disease progression, thereby enhancing overall effectiveness.

For intra-modality constraints, we apply similarity constraints to ensure the consistency of time-shared features within each modality, preserving the integrity of chest X-ray and diagnostic report characteristics over time. For inter-modality constraints, we implement multimodal contrastive constraint and multimodal structural constraint, respectively. The former is responsible for aligning time-shared and time-specific features of corresponding chest X-rays and reports while distancing non-matching pairs. This constraint also indirectly enhances the separation of features within the same modality, making the representation of time-shared and time-specific features more precise. The latter further regulates the spatial distribution of features by forming triangular structures, ensuring that the geometric relationships (e.g., distances and angles) within the triangles of image features correspond to those within text features. The combined effect of these constraints ensures that the generated reports more accurately reflect the progression of diseases, thereby enhancing their accuracy. 
Experimental results on the Longitudinal-MIMIC dataset demonstrate that our method achieves state-of-the-art performance on most NLG metrics, validating its effectiveness. Additionally, our method achieves superior results compared to other approaches without using historical information during testing and can be adapted to various multimodal large model frameworks, demonstrating strong applicability. We summarize contributions as follows:
\begin{itemize}
  \item We propose an innovative HC-LLM framework that leverages historical diagnostic data to improve the adaptability and performance of LLMs in RRG.
  
  \item We propose tri-consistency constraints that can effectively enhance the consistency and specificity of generated reports with historical data, ensuring alignment with disease progression observed in sequential chest X-rays.
  
  \item The proposed framework achieves superior performance without relying on historical data during testing and can be easily integrated with various multimodal large models, demonstrating its strong applicability. 
  
  \item Extensive evaluations on the Longitudinal-MIMIC dataset demonstrate that our method achieves state-of-the-art performance, underscoring its effectiveness and robustness in leveraging historical data to enhance radiology report generation with LLMs.
\end{itemize}

\begin{figure*}[t]
\centering
\includegraphics[width=0.85\linewidth]{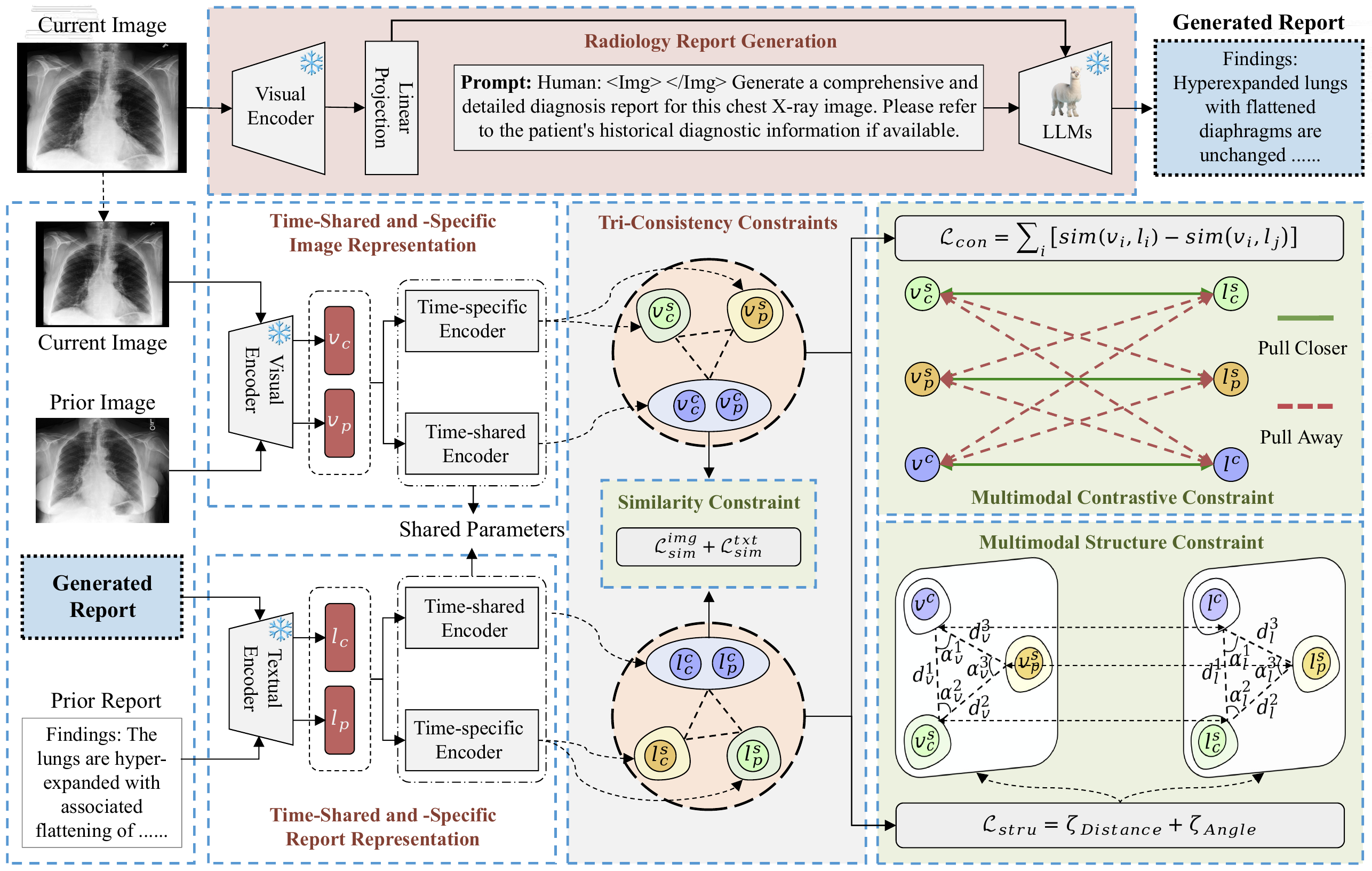}
\caption{\textbf{Overview of the proposed framework:} First, the current chest X-ray is processed to generate a diagnostic report using a visual encoder and LLM. The framework then extracts time-shared and time-specific features from the current and prior chest X-rays, along with the generated and prior diagnostic reports. Then, similarity constraints are first applied to ensure consistent time-shared representation over time. Finally, multimodal contrastive and structural constraints are employed to align shared and specific features across modalities, ensuring the generated report accurately reflects disease progression.}
\label{fig 2}
\vspace{-3mm}
\end{figure*}

\section{Related Works}
\textbf{Radiology Report Generation}: Radiology report generation methodologies have evolved significantly from early CNN-RNN frameworks \cite{vinyals2015show, lu2017knowing, wang2024ime, wang2024made} to the integration of advanced Transformer architectures \cite{li2022cross, wang2023metransformer, huang2023kiut}. Initial approaches focused on cross-modal alignment, utilizing CNNs for image features and RNNs for text generation. The advent of Transformers brought enhanced cross-modal interactions and long-range dependency modeling. Researchers introduced memory modules \cite{chen2020generating, chen2022cross, qin2022reinforced, cao2023mmtn, shen2024automatic},  hierarchical alignment \cite{you2021aligntransformer, li2023unify} and knowledge-guided enhancement techniques \cite{li2023auxiliary, huang2023kiut,li2023dynamic,li2024contrastive} to better capture multi-level interactions. Despite their advancements, these approaches are tailored for single chest X-ray report generation and are limited in processing multimodal inputs, which constrains their applicability in longitudinal radiology report generation.

Recently, the advent of LLMs has further brought significant advancements to RRG. For instance,  \citet{liu2024bootstrapping} proposed bootstrapping LLMs with in-domain instance induction and coarse-to-fine decoding to enhance alignment with medical data. \citet{jin2024promptmrg} proposed PromptMRG, which enhances radiology report generation by using diagnosis-driven prompts and addresses disease imbalance with adaptive loss techniques. Although LLMs can technically process sequential data by feeding the sequence directly into the model, this approach often encounters issues such as hallucination and suboptimal performance.  To overcome these challenges,  we propose a novel HC-LLM framework for RRG. By leveraging historical information and incorporating various types of constraints, HC-LLM ensures that LLM-generated reports can accurately capture the disease's progression nature.

\noindent
\textbf{Longitudinal Radiology Report Generation}: 
Recent advancements in RRG have increasingly focused on leveraging longitudinal information to improve the accuracy and relevance of generated reports. Current research can be divided into two main directions. The first direction involves using only historical chest X-ray information. For example, \citet{karwande2022chexrelnet} introduced an anatomy-aware model to track longitudinal relationships between chest X-rays, effectively capturing disease progression. \citet{bannur2023learning} proposed a self-supervised framework to model the longitudinal evolution of chest X-ray findings, enhancing the understanding of disease changes over time. Additionally, \citet{serra2023controllable} employed Faster R-CNN to create composite representations of longitudinal studies, highlighting anatomical changes. The second direction, which is more aligned with our focus, leverages both historical chest X-rays and diagnostic reports. This approach is crucial for capturing the full scope of disease progression and providing comprehensive context for current diagnoses. For instance, \citet{zhu2023utilizing} proposed a cross-attention-based multi-modal fusion framework to utilize patient record chronology, thereby improving report pre-filling tasks. Although these methods have made significant progress, they have not thoroughly explored the adaptation of LLMs to longitudinal medical data, often missing the complex progression of diseases, which significantly impacts the effectiveness of generated reports.

\section{Method}
\subsection{Problem Formulation}
\label{sec:method}
The overall framework of HC-LLM is illustrated in Figure~\ref{fig 2}. The input comprises the current chest X-ray image ($I_c$) and the previous chest X-ray image ($I_p$) along with its corresponding diagnostic report ($R_p$). The objective is to generate a diagnostic report ($\hat{R}_c$) for $I_c$, that closely approximates the ground truth report ($R_c$). Formally, given:
\begin{equation}
\hat{R}_c \xleftarrow{} \text{HC-LLM}(I_c, (I_p, R_p)).
\end{equation}
Notably, our framework is flexible and supports two testing scenarios: 1) leveraging historical diagnostic information; 2) relying solely on the current chest X-ray image. Additionally, it can be extended to incorporate diagnostic information from multiple historical time points to assist in generating the current report. This flexibility enhances the practical applicability of our model in real-world clinical settings.

\subsection{Radiology Report Generation} 
The overall workflow of the RRG is illustrated at the top part of Figure~\ref{fig 2}. This process consists of three main components: visual encoding, prompt generation, and report generation with the LLMs. Firstly, the visual encoder processes the current chest X-ray image $I_c$ using the Swin Transformer \cite{liu2021swin} $f_{ve}(\cdot)$, extracting latent visual features that capture the anatomical and pathological details from the radiograph. Formally, the visual feature extraction is defined as:
\begin{equation}
    f_{ve}(I_{c}) = X = \{x_1, x_2, \ldots, x_S\},
\end{equation}
where $x_i \in \mathbb{R}^d$ is a feature patch, $d$ denotes the feature dimension, and $S$ is the number of patches. Next, for prompt generation, we define a general prompt $p_g$ as shown in the middle section of the top part of Figure~\ref{fig 2}. It is important to note that the prompt can be adapted to include or exclude historical diagnostic information based on the input provided during testing. Finally, the radiology report generation component utilizes a large language model $f_{tg}$ to produce the diagnostic report $\hat{R}{c}$. Each report is represented as $\hat{R}_{c} = \{\hat{r}_1, \hat{r}_2, \ldots, \hat{r}_T\}$, where $\hat{r}_i \in \mathbb{V}$ is a token, $T$ is the length of the report, and $\mathbb{V}$ represents the vocabulary. The decoding process is formulated as:
\begin{equation}
    \hat{r}_t = f_{tg}(X, p_g, \hat{r}_{1:t-1}),
\end{equation}
where $\hat{r}_t$ is the token to be predicted at token step $t$. The model is optimized based on the cross-entropy loss $\mathcal{L}_{RRG}$ from the final generated reports $\hat{R}_{c}$ and the gold standard $R_{c}$. The primary loss function is defined as:
\begin{equation}
    \mathcal{L}_{RRG} = -\sum_{t=1}^{T} \log p(\hat{r}_t | \hat{r}_{1:t-1}, X, p_{g}).
\end{equation}

\subsection{Time-Shared and -Specific Representations}
For chest X-rays and reports at two different time points, diseases often exhibit characteristics such as disappearance, stability, and emergence within each respective modality space. Therefore, we construct both time-shared and specific features for the chest X-rays and reports to capture these characteristics.  For the images, we use the previously mentioned visual encoder $f_{ve}$ and take the output at the \texttt{CLS} position as the representations of the two images, followed by a linear mapping layer to project these features into the text space, as shown below:
\begin{equation}
v_{c}, v_{p} = W_v \cdot f_{ve}(I_c)[\texttt{CLS}], W_v \cdot f_{ve}(I_p)[\texttt{CLS}],
\end{equation}
where $W_{v}$ represents the linear mapping layer for visual features. For the generated and historical reports, we separately input them into the text encoder 
$f_{tg}$ and also take the output at the \texttt{CLS} position as their respective representations, as shown below:
\begin{equation}
l_{c}, l_{p} = f_{tg}(\hat{R}_{c})[\texttt{CLS}], f_{tg}(R_{p})[\texttt{CLS}].
\end{equation}

The time-shared and specific features are then extracted using dedicated encoders for both image and text modalities, as described below:
\begin{equation}
\begin{gathered}
y_{c}^{c}, y_{c}^{s} = E_{c}(LN(y_{c}), \theta^{c}), E_{c}^{s}(LN(y_{c}), \theta_{c}^{s}), y \in \{v,l\}\\
y_{p}^{c}, y_{p}^{s} = E_{c}(LN(y_{p}), \theta^{c}), E_{p}^{s}(LN(y_{p}), \theta_{p}^{s}), y \in \{v,l\},
\end{gathered}
\end{equation}
where $y$ represents the modality, with $y_c^c$, $y_p^c$ as shared features and $y_c^s$, $y_p^s$ as specific features at current and prior times. $LN(\cdot)$ denotes the Layer Normalization. $E_{c}(\cdot)$ represents the shared encoder, while $E_{c}^{s}(\cdot)$ and $E_{p}^{s}(\cdot)$ represent the time-specific encoders for current and previous data, respectively. Specifically, using the same set of encoders for both image and text modalities could enhance multimodal alignment and integration, enabling more effective constraint application across modalities.

\begin{table*}[!tb]\small 
\centering
\setlength\tabcolsep{0.75mm} 
\renewcommand{\arraystretch}{0.9}
\begin{tabular}{c|l|c|c|cccccc|ccc} \hline
\multirow{2}{*}{\textbf{Dataset}} & \multirow{2}{*}{\textbf{Model}} & \multirow{2}{*}{\textbf{Year}} & \multirow{2}{*}{\textbf{Inputs}} & \multicolumn{6}{c|}{\textbf{NLG metrics}} & \multicolumn{3}{c}{\textbf{CE metrics}}
\\ \cline{5-13} & & & & BLEU-1 & BLEU-2 & BLEU-3 & BLEU-4 & METEOR & ROUGE-L & PREC & REC & F-1
\\ \hline 
\multirow{22}{*}{\textbf{\makecell{Longitudinal\\-MIMIC}}} & AoANet & 2019 & Single & 0.272 & 0.168 & 0.112 & 0.080 & 0.115 & 0.249 & 0.437  & 0.249 & 0.317\\
& CNN+Trans & 2019 & Single & 0.299 & 0.186 & 0.124 & 0.088 & 0.120 & 0.263 & 0.445 & 0.258 & 0.326\\
& R2Gen & 2020 & Single & 0.302 & 0.183 & 0.122 & 0.087 & 0.124 & 0.259 & 0.500 & 0.305 & 0.379\\ 
& R2CMN & 2021 & Single & 0.305 & 0.184 & 0.122 & 0.085 & 0.126 & 0.265 & 0.521 & 0.396 & 0.449\\
& R2GenRL & 2022 & Single & 0.303 & 0.153 & 0.082 & \underline{0.136} & - & 0.175 & 0.435 & 0.464 & 0.419\\
& CvT2DistilGPT2 & 2023 & Single & 0.365 & 0.226 & 0.151 & 0.107 & 0.143 & 0.275 & 0.443 & 0.369 & 0.379\\ 
& PromptMRG & 2024 & Single & 0.370 & 0.219 & 0.141 & 0.098 & 0.144 & 0.266 & 0.519 & \textbf{0.507} & \underline{0.482}\\
& Prefilling & 2023 & Longitudinal & 0.343 & 0.210 & 0.140 & 0.099 & 0.137 & 0.271 & \underline{0.538} & 0.434 & 0.480\\ 
\cline{2-13}
& R2GenGPT\ding{171} & 2023 & Single & 0.358 & 0.224 & 0.150 & 0.103 & 0.235 & 0.269 & 0.228 & 0.151 & 0.168\\
& + report & 2023 & Longitudinal & 0.367 & 0.223 & 0.145 & 0.100 & 0.124 & 0.265 & 0.460 & 0.435 & 0.416\\
& + image & 2023 & Longitudinal & 0.332 & 0.194 & 0.125 & 0.082 & 0.145 & 0.237 & 0.341 & 0.267 & 0.277\\
& + report \& image & 2023 & Longitudinal & 0.389 & 0.246 & 0.166 & 0.117 & 0.228 & 0.278 & 0.402 & 0.358 & 0.352\\
& \textbf{HC-LLM(Ours)\ding{171}} & - & Longitudinal & 0.404 & \underline{0.260} & \underline{0.178} & 0.128 & 0.160 & \textbf{0.287} & 0.417 & 0.357 & 0.357\\
\cline{2-13}
& BioMedGPT\ding{169} & 2023 & Single & 0.365 & 0.230 & 0.155 & 0.111 & 0.085 & 0.266 & 0.269 & 0.242 & 0.237\\
& + report & 2023 & Longitudinal & 0.393 & 0.252 & 0.172 & 0.121 & \underline{0.232} & 0.281 & 0.381 & 0.314 & 0.321\\
& + image & 2023 & Longitudinal & 0.356 & 0.225 & 0.149 & 0.102 & 0.133 & 0.265 & 0.252 & 0.176 & 0.194\\
& + report \& image & 2023 & Longitudinal & 0.398 & 0.254 & 0.173 & 0.121 & \textbf{0.279} & 0.281 & 0.401 & 0.340 & 0.341\\
& \textbf{HC-LLM(Ours)\ding{169}} & - & Longitudinal & \underline{0.406} & \underline{0.260} & \underline{0.178} & 0.127 & 0.162 & \underline{0.285} & 0.415 & 0.358 & 0.360\\ 
\cline{2-13}
& MiniGPT4\ding{168} & 2023 & Single & 0.375 & 0.231 & 0.150 & 0.099 & 0.135 & 0.266 & 0.193 & 0.112 & 0.133\\
& + report & 2023 & Longitudinal & 0.405 & 0.255 & 0.172 & 0.119 & 0.156 & 0.281 & 0.420 & 0.374 & 0.366\\
& + image & 2023 & Longitudinal & 0.365 & 0.226 & 0.149 & 0.100 & 0.146 & 0.266 & 0.182 & 0.129 & 0.141\\
& + report \& image & 2023 & Longitudinal & 0.395 & 0.248 & 0.166 & 0.114 & 0.144 & 0.280 & 0.411 & 0.343 & 0.346\\
& \textbf{HC-LLM(Ours)\ding{168}} & - & Longitudinal & \textbf{0.416} & \textbf{0.276} & \textbf{0.193} & \textbf{0.142} & 0.162 & 0.284 & \textbf{0.617} & \underline{0.494} & \textbf{0.498}\\ 
\hline
\end{tabular}
\caption{Results of the NLG metrics (BLEU (BL), Meteor (MTR), Rouge-L (R-L)) and clinical efficacy (CE) metrics (Precision (PREC), Recall (REC) and F-1 score) on the \textit{Longitudinal-MIMIC} dataset. Best results are highlighted in bold, and the second best are underlined. Identical symbols (i.e., \ding{171}, \ding{169}, \ding{168}) in the table denote models using the same architecture.}
\label{table 1}  
\vspace{-0.4cm}
\end{table*}

\subsection{Tri-Consistency Constraints}
Based on the time-shared and specific features of longitudinal chest X-rays and reports, we further introduce three constraints to enhance the performance of the LLMs in generating medical reports, ensuring that the generated reports accurately reflect the disease progression characteristics.

\noindent
\textbf{Similarity Constraint $\mathcal{L}_{sim}$.} The similarity constraint is designed to align the time-shared features within each modality. Among various metric choices, we employ the Mean Squared Error (MSE) for this purpose, which measures the discrepancy by computing the average of the squares of the differences between corresponding values. The MSE loss is defined as follows:
\begin{equation}
\mathcal{L}_{sim}^{img} = \frac{1}{2} \sum \|v_c^c - v_p^c\|_2^2,\quad\mathcal{L}_{sim}^{txt} = \frac{1}{2} \sum \|l_c^c - l_p^c\|_2^2.
\end{equation}

Despite its sensitivity to outliers, its simplicity and computational efficiency make it ideal for our similarity loss, enhancing the consistency of time-shared representations across different time points.

\noindent
\textbf{Multimodal Contrastive Constraint $\mathcal{L}_{con}$.} The introduction of $\mathcal{L}_{con}$ serves two main purposes. Firstly, it aligns the shared and specific features of corresponding chest X-rays and reports, ensuring that the progression characteristics of diseases in images are consistent with those in reports. Secondly, by bringing the matching features between images and reports closer and distancing the non-matching features, it indirectly promotes the separation of the three features within the same modality, thereby enhancing their specific semantic information. Considering the semantic consistency of shared features and to facilitate the implementation of contrastive constraint, we perform average pooling on the two shared features within the same modality, as follows:
\begin{equation}
y^{c} = (y_c^{c} + y_p^{c})/2, \quad y \in \{v, l\}.
\end{equation}

Then, for the image sequence $\tilde{v} = [v^{c}, v_{p}^{s}, v_{c}^{s}]$ and text sequence $\tilde{l} = [l^{c}, l_{p}^{s}, l_{c}^{s}]$, we use the InfoNCE loss $\mathcal{L}_{con}$, which includes an image-to-text contrastive loss $\mathcal{L}_{i2t}$ and a text-to-image contrastive loss $\mathcal{L}_{t2i}$ to achieve the aforementioned objectives, denoted as:
\begin{equation}
\mathcal{L}_{con} = (\mathcal{L}_{i2t} + \mathcal{L}_{t2i})/2,
\end{equation}
where the image-to-text contrastive loss $\mathcal{L}_{i2t}$ is formulated as:
\begin{equation}
\mathcal{L}_{i2t} = - \log \frac{\exp((\tilde{v}_i, \tilde{l}_i) / \tau)}{\sum_{k=1}^{3} \exp((\tilde{v}_i, \tilde{l}_k) / \tau)},
\end{equation}
where $\tau$ is the temperature hyper-parameter. Similarly, the text-to-image contrastive loss 
$\mathcal{L}_{t2i}$ is
\begin{equation}
\mathcal{L}_{t2i} = - \log \frac{\exp((\tilde{l}_i, \tilde{v}_i) / \tau)}{\sum_{k=1}^{3} \exp((\tilde{l}_i, \tilde{v}_k) / \tau)}.
\end{equation}

By aligning the evolution of disease characteristics between images and text, we ensure that the generated reports more accurately reflect the longitudinal progression of medical conditions.

\noindent
\textbf{Multimodal Structural Constraint $\mathcal{L}_{stru}$.} 
While multimodal contrastive constraint effectively pulls together matched features and pushes apart unmatched features, it does not sufficiently guarantee consistent structural relationships in the feature space. Thus, we further introduce multimodal structural constraint $\mathcal{L}_{stru}$, which ensures that the geometric relationships among features from images and reports remain consistent in the feature space. Following the methodology proposed by \citet{park2019relational}, we define the structural loss $\mathcal{L}_{stru}$ as a combination of distance-wise and angle-wise constraints to enhance the structural consistency of the representations.

\textbf{Distance-wise loss:} This constraint aligns the distances between features in the image space with those in the text space. Given a pair of feature representations ($t_{i},t_{j}),t \in \{\tilde{v},\tilde{l}\}$, the distance-wise function $\psi_D$ measures the Euclidean distance between the two features as follows:
\begin{equation}
    \psi_D(t_i, t_j) = \frac{1}{\mu} \| t_i - t_j \|_2,
\end{equation}
where $\mu$ is a normalization factor for distance, defined as the average distance between pairs $\mathcal{X}^{2}$ within that modality:
\begin{equation}
    \mu = \frac{1}{|\mathcal{X}^2|} \sum_{(t_i, t_j) \in \mathcal{X}^2} \| t_i - t_j \|_2.
\end{equation}

The distance-wise constraint loss is defined as follows:
\begin{equation}
\mathcal{L}_{distance} = \sum_{\substack{i,j=1, i \neq j}}^3 l_\delta(\psi_D(\tilde{v}_i, \tilde{v}_j), \psi_D(\tilde{l}_i, \tilde{l}_j)),
\end{equation}
where $l_\delta$ is the Huber loss, defined as:
\begin{equation}
    l_\delta(x, y) = 
    \begin{cases} 
        \frac{1}{2}(x - y)^2, & \text{if } |x - y| \leq 1, \\
        |x - y| - \frac{1}{2}, & \text{otherwise}.
    \end{cases}
\end{equation}

\textbf{Angle-wise loss:} 
This loss ensures that the angles between triplets of features in the image modality are consistent with those in the text modality. This is achieved by calculating the angles formed by three examples using cosine similarity:
\begin{equation}
    \psi_A(t_i, t_j, t_k) = \frac{\langle t_i - t_j, t_i - t_k \rangle}{\| t_i - t_j \| \cdot \| t_i - t_k \|},t \in \{\tilde{v},\tilde{l}\}.
\end{equation}

The angle-wise constraint is defined as follows:
\begin{equation}
    \mathcal{L}_{angle} = \sum_{\substack{i, j, k \in \{1,2,3\} \\ i \neq j \neq k}} l_\delta (\psi_A(\tilde{v}_i, \tilde{v}_j, \tilde{v}_k), \psi_A(\tilde{l}_i, \tilde{l}_j, \tilde{l}_k)).
\end{equation}

Thus, the final $\mathcal{L}_{stru}$ can be summarized as:
\begin{equation}
\mathcal{L}_{stru} = \mathcal{L}_{distance} + \mathcal{L}_{angle}.
\end{equation}

By minimizing this structural loss, we ensure that the geometric relationships within the image features are mirrored in the text features, thereby reinforcing the structural consistency between both modalities.

\begin{figure*}[!ht]
\centering
\includegraphics[width=0.95\linewidth]{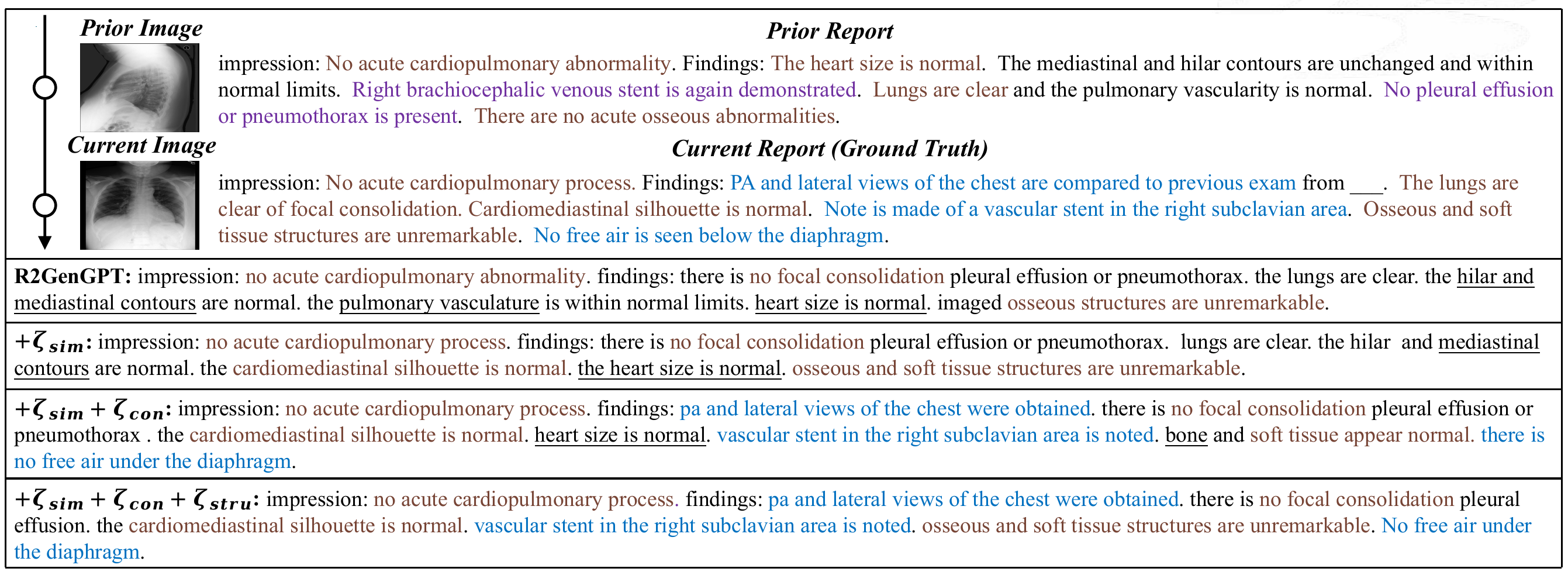}
\vspace{-2mm}
\caption{An illustration of reports generated by different models using longitudinal images and the historical report. Brown denotes common content, while purple and blue indicate time-specific content. Underlined text marks incorrect predictions.}
\label{fig 3}
\end{figure*}

\begin{table*}[!tb]\small 
\centering
\setlength\tabcolsep{2.4mm} 
\renewcommand{\arraystretch}{1.0}
\begin{tabular}{ccc|cccccc|ccc} \hline
\multirow{2}{*}{$\mathcal{L}_{sim}$} & \multirow{2}{*}{$\mathcal{L}_{con}$} & \multirow{2}{*}{$\mathcal{L}_{stru}$} & \multicolumn{6}{c|}{NLG metrics} & \multicolumn{3}{c}{CE metrics}
\\ \cline{4-12}
& & & BLEU-1 & BLEU-2 & BLEU-3 & BLEU-4 & METEOR & ROUGE-L & PREC & REC & F-1
\\ \hline 
\ding{55} & \ding{55} & \ding{55} & 0.389 & 0.246 & 0.166 & 0.117 & 0.228 & 0.278 & 0.402 & 0.358 & 0.352 \\
 \ding{55} & \ding{51} & \ding{51} &  0.403 & 0.255 & 0.172 & 0.119 & \textbf{0.232} & 0.282 & \textbf{0.419}  & \textbf{0.362} & \textbf{0.361}\\
\ding{51} & \ding{55} & \ding{51} & 0.383 & 0.244 & 0.166 & 0.118 & 0.148 & 0.278 & 0.367 & 0.311 & 0.311\\	
\ding{51} & \ding{51} &  \ding{55} & 0.399 & 0.252 & 0.171 & 0.121 & 0.230 & 0.280 & 0.417 & 0.355 & 0.356\\	
\ding{51} & \ding{51} & \ding{51} & \textbf{0.404} & \textbf{0.260} & \textbf{0.178} & \textbf{0.128} & 
0.160 & \textbf{0.287} & 0.417 & 0.357 & 0.357\\	
\hline
\end{tabular}
\caption{Ablation study of each constraint on the dataset of \textit{Longitudinal-MIMIC}.}
\label{table 2} 
\vspace{-0.3cm}
\end{table*}

\subsection{Learning Objective}
The overall learning of the model is performed by minimizing:
\begin{equation}
\mathcal{L}_{total} = \mathcal{L}_{RRG} + \beta_1 ( \mathcal{L}_{sim}^{img} + \mathcal{L}_{sim}^{txt}) + \beta_2 \mathcal{L}_{con} + \beta_3 \mathcal{L}_{stru},
\end{equation}
where $\beta_1$, $\beta_2$, $\beta_3$ are the hyperparameters that determine the contribution of each regularization component to the overall loss $\mathcal{L}_{total}$.

\section{Experiments}
\noindent \textbf{Dataset:} Building on the dataset presented in \cite{zhu2023utilizing}, we utilized the Longitudinal-MIMIC dataset, which is derived from MIMIC-CXR, for our evaluation. This dataset was constructed by selecting patients with at least two visit records, resulting in a comprehensive dataset of 26,625 patients and a total of 94,169 samples. Each sample used for model training included the current visit's chest X-ray (CXR) and report, as well as the previous visit's CXR and report. The dataset was divided into training (26,156 patients and 92,374 samples), validation (203 patients and 737 samples), and test (266 patients and 2,058 samples) sets. More Details of datasets can be referred to Appendix.

\noindent \textbf{Evaluation Metrics:} We assess the performance of our model using both natural language generation (NLG) metrics and clinical efficacy (CE) metrics. For NLG, we utilize the BLEU \cite{papineni2002bleu}, METEOR \cite{denkowski2011meteor}, and ROUGE-L \cite{lin2004rouge} metrics. In accordance with the methodology proposed by \citet{nicolson2023improving}, our CE evaluation involves precision, recall, and F1 scores. These metrics are derived by converting generated reports into 14 disease classification labels using CheXbert \cite{smit2020chexbert}.

\noindent \textbf{Implementation Details:} In this study, for the R2GenGPT framework, we selected the base version of the Swin Transformer\footnote{\url{https://huggingface.co/microsoft/swin-base-patch4-window7-224}} as the visual encoder and used the LLAMA2-7B\footnote{\url{https://huggingface.co/meta-llama/Llama-2-7b-chat-hf}} model as the primary language model for both R2GenGPT and MiniGPT4 frameworks.  BioMedGPT\footnote{\url{https://huggingface.co/PharMolix/BioMedGPT-LM-7B}} maintains consistency with the R2GenGPT image encoder and utilizes BioMedGPT-LM-7B as its language model. The coefficients were set to $\beta_1 = 1.0$, $\beta_2 = 0.8$, and $\beta_3 = 1.0$, respectively. The training process was executed on a single NVIDIA A800 80GB GPU using mixed precision for 5 epochs on the Longitudinal-MIMIC dataset, with a mini-batch size of 4 and a learning rate of 1e-4. For the testing phase, we employed a beam search strategy with a beam size of 3.
 
\subsection{Comparison with State-of-the-Art Methods}
We compared our method with single time-point RRG methods (i.e., AoANet \cite{huang2019attention}, CNN+Trans, R2Gen \cite{chen2020generating}, R2CMN \cite{chen2022cross}, R2GenRL \cite{qin2022reinforced}, CvT2DistilGPT2 \cite{nicolson2023improving}, PromptMRG \cite{jin2024promptmrg}) and longitudinal RRG methods (i.e., Prefilling \cite{zhu2023utilizing}, R2GenGPT \cite{wang2023r2gengpt}, BioMedGPT \cite{luo2023biomedgpt}, MiniGPT4 \cite{zhu2023minigpt}). As shown in Table~\ref{table 1}, our method achieves improvements in most metrics. Specifically, compared to single time-point methods, longitudinal models generally exhibit superior performance, demonstrating the importance of modeling longitudinal information for the RRG task. Additionally, for longitudinal models, traditional cross-attention methods fall short of LLM-based approaches due to the latter's superior semantic modeling capabilities. Nevertheless, LLM-based baselines still underperform compared to our proposed method, primarily because simply inputting longitudinal data into large language models does not fully utilize the unique longitudinal characteristics. In contrast, our model constrains the consistency of disease progression within longitudinal chest X-rays and reports, ensuring that the LLM-generated reports accurately reflect disease progression, thereby enhancing their accuracy. Additionally, we can observe that our method achieves performance improvements across different frameworks, demonstrating its strong applicability. 
More experimental results and analyses are available in the Appendix.

\subsection{Model Analysis}
\noindent
\textbf{Ablation Study.}
Table~\ref{table 2} presents an ablation study of each constraint on the Longitudinal-MIMIC dataset. Firstly, when the similarity constraint $\mathcal{L}_{sim}$ is removed, there is only a minor performance drop. This can be attributed to the fact that although removing the similarity constraint disrupts the alignment of shared features, the presence of contrastive and structural constraints still maintains a certain degree of cross-modal longitudinal consistency. Secondly, removing the contrastive constraint $\mathcal{L}_{con}$ results in a significant performance decrease. The core reason is that the lack of a contrastive constraint hinders the alignment of shared and specific features across modalities. Additionally, this indirectly prevents the promotion of separation between features within the same modality, thereby failing to ensure the accuracy of specific features, which ultimately impacts the overall model performance. Lastly, when the structural constraint $\mathcal{L}_{stru}$ is removed, there is also a noticeable performance decline, highlighting its importance in maintaining cross-modal longitudinal consistency and ensuring the effectiveness of the LLMs' outputs.

\begin{figure}[t]
\centering
\includegraphics[width=1.0\linewidth]{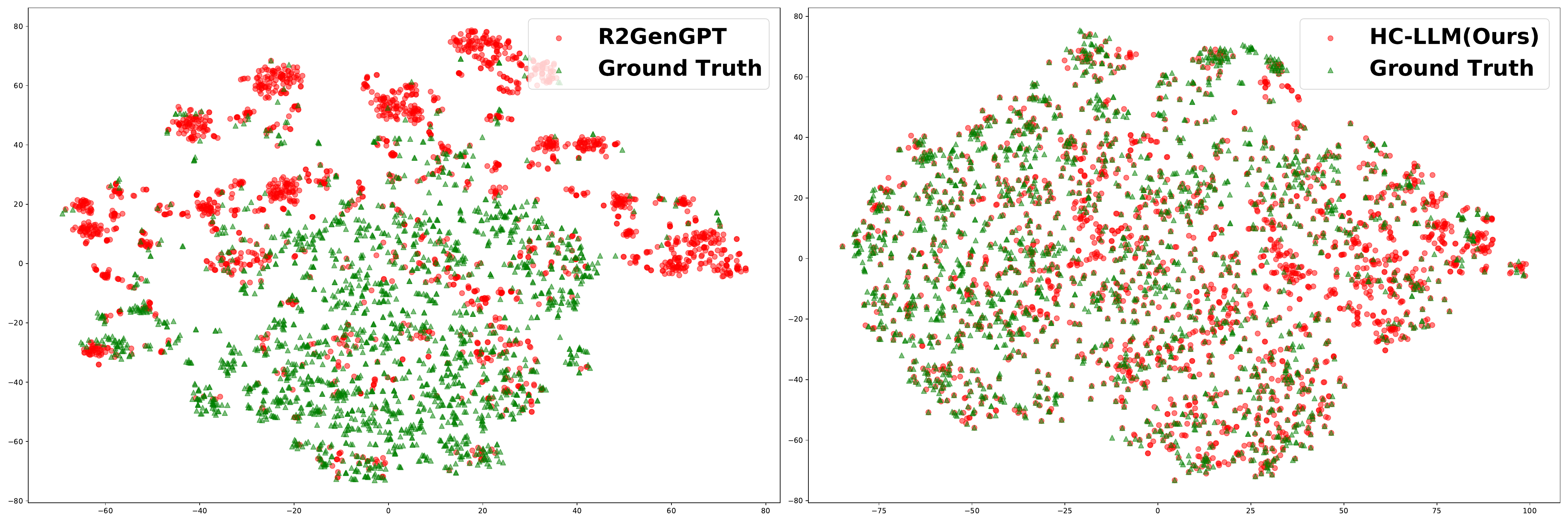}
\vspace{-5mm}
\caption{Visualization of feature distributions using t-SNE for the R2GenGPT and HC-LLM (Ours) models.}
\label{fig 5}
\end{figure}

\begin{table}[t]
\centering
\setlength\tabcolsep{1.0mm} 
\renewcommand{\arraystretch}{0.9}
\begin{tabular}{lccccccc}
\hline
Models & BL-1 & BL-2 & BL-3 & BL-4 & MTR & R-L \\
\hline
PromptMRG & 0.370 & 0.219 & 0.141 & 0.098 & 0.144 & 0.266\\
Prefilling & 0.253 & 0.159 & 0.107 & 0.077 & 0.118 & \textbf{0.269}\\
R2GenGPT & 0.358 & 0.224 & 0.150 & 0.103 & \textbf{0.235} & \textbf{0.269}\\
BioMedGPT & 0.346 & 0.211 & 0.136 & 0.088 & 0.096 & 0.255\\
MiniGPT4 & 0.344 & 0.181 & 0.106 &0.063 & - & 0.222\\
HC-LLM(Ours) & \textbf{0.371} & \textbf{0.231} & \textbf{0.154} & \textbf{0.107} & 0.127 & 0.268\\
\hline
\end{tabular}
\caption{Performance comparison without historical information during testing. The HC-LLM model operates within the R2GenGPT framework.}
\label{table 3}
\vspace{-0.5cm}
\end{table}

\noindent
\textbf{Qualitative results.}
To qualitatively demonstrate how historical information, under different constraints, better adapts LLMs to RRG, we perform a case study on the output reports generated with various combinations of constraints using the same longitudinal inputs. As shown in Figure~\ref{fig 3}, the reports at different time points indeed show disease disappearance (purple), stability (brown), and new occurrences (blue). When feeding longitudinal chest X-rays and historical reports directly to R2GenGPT, it generates more generic content, not fully utilizing longitudinal data to produce a targeted report for the current X-ray.  Upon introducing similarity constraints, the generated report includes more common content, aligning with our expected outcome. With the addition of contrastive constraint, both common and unique contents are reflected in the generated report. This is mainly due to the contrastive constraint ensuring that the distinctive features in the generated report align with those in the current chest X-ray, effectively promoting the generation of specific content. Finally, by introducing structural constraint, we observe that the generated report's accuracy improves, and certain erroneous predictions are eliminated. The contrastive constraint helps maintain consistency, while the structural constraint significantly enhances this consistency, providing better regulation and adaptation to the LLMs' generative performance. To more intuitively display the distribution of features before and after applying constraints, we used t-SNE to visualize the distributions between the R2GenGPT, our method, and the actual reports. As shown in Figure~\ref{fig 5}, the distribution of our method aligns more closely with the actual reports, more directly confirming its superior generative performance.

\noindent
\textbf{Testing Performance without Historical Data.} We further evaluate the longitudinal models using only the current chest X-ray during testing. As shown in Table~\ref{table 3}, the performance of traditional models drops significantly, demonstrating their limited applicability. While methods based on LLMs are relatively more stable, they also experience some performance decline. Notably, R2GenGPT, BioMedGPT and MiniGPT4 perform worse than single time-point PromptMRG when tested using only the current chest X-ray. This is primarily because it does not effectively utilize historical information to adapt LLMs to the RRG task. In contrast, our model outperforms the PromptMRG method. This superior performance is attributed to our training process, which better captures the evolutionary characteristics of diseases in longitudinal data and more effectively adapts LLMs to RRG.

\begin{figure}[t]
\centering
\includegraphics[width=1.0\linewidth]{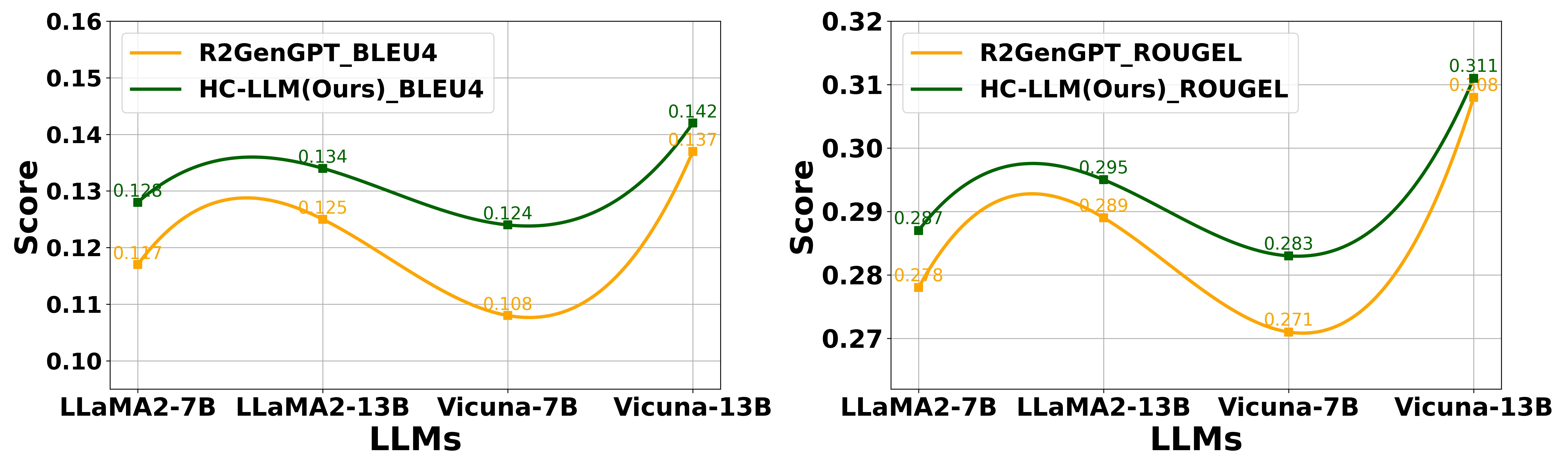}
\vspace{-5mm}
\caption{Performance comparison of BLEU-4 and ROUGE-L scores for R2GenGPT and HC-LLM(Ours) models across different LLMs.}
\label{fig 4}
\vspace{-0.4cm}
\end{figure}

\noindent
\textbf{Performance Analysis Under Different LLMs.} Figure~\ref{fig 4} shows the results of our method under different LLMs. As observed, regardless of whether LLaMA2-7B, LLaMA2-13B, Vicuna-7B, or Vicuna-13B is used, our method consistently achieves better results in both BLEU-4 and ROUGE-L metrics. This indicates that our model has good adaptability and robustness across various LLM architectures, leading to stable improvements in report generation quality. Notably, larger models exhibit improved performance, likely because they contain more general information and can better adapt to RRG with the help of our introduced constraints.

\section{Conclusion}
In this paper, we propose a novel HC-LLM framework that leverages historical diagnostic information to ensure that the reports generated by LLMs better align with the progression of diseases. Experimental results demonstrate that our method exhibits superior performance with both single chest X-ray data and longitudinal data during testing, proving its effectiveness. Additionally, our architecture can easily adapt to different multimodal large model frameworks and achieve substantial performance improvements, demonstrating its excellent applicability. This method provides a practical paradigm for adapting general LLMs to sequential data applications. Currently, HC-LLM only uses two-time point longitudinal data and has not yet explored more complex diagnostic data from multiple historical time points, which is key for understanding disease progression and could be explored in the future to further improve performance.

\section{Acknowledgments}
This work was supported by National Key R\&D Program of China No.2021ZD0111902 and National Natural Science Foundation of China under Grant 62172022, Grant U21B2038, and Grant 62476179.

\bibliography{aaai25}

\appendix

\section{Appendix}

\subsection{Datasets} 

\noindent
\textbf{MS-CXR-T Dataset:} 
This dataset was originally designed for image classification and sentence similarity tasks.  The former comprises
multi-image and ground-truth label pairs $(N = 1326)$ across
5 findings, with classes corresponding to 3 states of disease
progression for each finding: {Improving, Stable,
Worsening}. To validate our model's performance in generating longitudinal reports, we extracted data from the MIMIC dataset according to the indices of chest X-rays and reports provided by the dataset, forming the MS-CXR-T dataset suitable for longitudinal RRG. Considering the number of samples, we divided the data into a training set and a test set in an 8:2 ratio.

\begin{table*}[!tb]\small 
\centering
\setlength\tabcolsep{0.8mm} 
\renewcommand{\arraystretch}{1.0}
\begin{tabular}{c|l|c|c|cccccc|ccc} \hline
\multirow{2}{*}{\textbf{Dataset}} & \multirow{2}{*}{\textbf{Model}} & \multirow{2}{*}{\textbf{Year}} & \multirow{2}{*}{\textbf{Inputs}} & \multicolumn{6}{c|}{\textbf{NLG metrics}} & \multicolumn{3}{c}{\textbf{CE metrics}}
\\ \cline{5-13} & & & & BLEU-1 & BLEU-2 & BLEU-3 & BLEU-4 & METEOR & ROUGE-L & PREC & REC & F-1
\\ \hline 
\multirow{20}{*}{\textbf{MS-CXR-T}} & R2Gen & 2020 & Single & 0.139 & 0.090 & 0.066 & 0.052 & 0.133 & 0.138 & 0.189 & 0.215 & 0.180\\ 
& R2CMN & 2021 & Single & 0.186 & 0.111 & 0.074 & 0.053 & 0.086 & 0.211 & 0.162 & 0.316 & 0.196\\ 
& R2GenRL & 2022 & Single & 0.328 & 0.179 & 0.109 & 0.075 & - & 0.211 & 0.396 & 0.356 & 0.349\\ 
& CvT2DistilGPT2 & 2023 & Single & 0.334 & 0.195 & 0.136 & 0.105 & 0.149 & 0.254 & 0.247 & 0.234 & 0.240\\  
& Prefilling & 2023 & Longitudinal & 0.214 & 0.096 & 0.054 & 0.032 & 0.106 & 0.192 & 0.300 & 0.308 & 0.277\\ 
\cline{2-13}
& R2GenGPT\ding{171} & 2023 & Single & 0.335 & 0.197 & 0.133 & 0.100 & 0.186 & 0.259 & 0.295 & 0.171 & 0.204\\
& + report & 2023 & Longitudinal & 0.351 & 0.228 & 0.165 & 0.129 & 0.126 & 0.285 & 0.539 & 0.504 & 0.485\\
& + image & 2023 & Longitudinal & 0.329 & 0.195 & 0.130 & 0.097 & 0.164 & 0.256 & 0.242 & 0.157 & 0.178\\
& + report \& image & 2023 & Longitudinal & 0.356 & 0.230 & 0.165 & 0.128 & 0.169 & 0.286 & 0.486 & 0.455 & 0.433\\ 
& \textbf{HC-LLM(Ours)\ding{171}} & - & Longitudinal & \underline{0.364} & \textbf{0.238} & \textbf{0.174} & \textbf{0.136} & \textbf{0.231} & \textbf{0.296} & 0.515 & 0.492 & 0.474\\
\cline{2-13}
& BiomedGPT\ding{169} & 2023 & Single & 0.323 & 0.190 & 0.128 & 0.096 & 0.166 & 0.254 & 0.290 & 0.189 & 0.209\\
& + report & 2023 & Longitudinal & 0.354 & 0.230 & 0.167 & 0.129 & 0.110 & 0.287 & 0.532 & 0.509 & 0.481\\
& + image & 2023 & Longitudinal & 0.330 & 0.199 & 0.137 & 0.104 & 0.083 & 0.263 & 0.246 & 0.225 & 0.217\\
& + report \& image & 2023 & Longitudinal & 0.355 & 0.230 & 0.167 & 0.130 & 0.218 & 0.287 & 0.535 & \underline{0.513} & 0.490\\
& \textbf{HC-LLM(Ours)\ding{169}} & - & Longitudinal & \textbf{0.366} & \textbf{0.238} & \underline{0.172} & \underline{0.134} & 0.210 & \underline{0.291} & 0.464 & 0.448 & 0.423\\ 
\cline{2-13}
& MiniGPT4\ding{168} & 2023 & Single & 0.318 & 0.182 & 0.121 & 0.090 & 0.145 & 0.244 & 0.243 & 0.190 & 0.199\\
& + report & 2023 & Longitudinal & 0.334 & 0.216 & 0.157 & 0.123 & 0.214 & 0.295 & 0.508 & 0.476 & 0.459\\
& + image & 2023 & Longitudinal & 0.308 & 0.177 & 0.112 & 0.083 & 0.149 & 0.242 & 0.204 & 0.138 & 0.154\\
& + report \& image & 2023 & Longitudinal & 0.347 & 0.225 & 0.163 & 0.128 & 0.218 & 0.282 & \underline{0.542} & 0.507 & \underline{0.491}\\
& \textbf{HC-LLM(Ours)\ding{168}} & - & Longitudinal & 0.358 & \underline{0.233} & 0.170 & 0.133 & \underline{0.221} & 0.283 & \textbf{0.567} & \textbf{0.528} & \textbf{0.514}\\ 
\hline
\end{tabular}
\caption{Results of the NLG metrics (BLEU (BL), Meteor (MTR), Rouge-L (R-L)) and clinical efficacy (CE) metrics (Precision (PREC), Recall (REC) and F-1 score) on the \textit{MS-CXR-T} dataset. Best results are highlighted in bold, and the second best are underlined. Identical symbols (i.e., \ding{171}, \ding{169}, \ding{168}) in the table denote models using the same architecture.}
\label{table 4}   
\end{table*}

\subsection{Experiment results on MS-CXR-T}
\begin{figure}[t]
\centering
\includegraphics[width=1.0\linewidth]{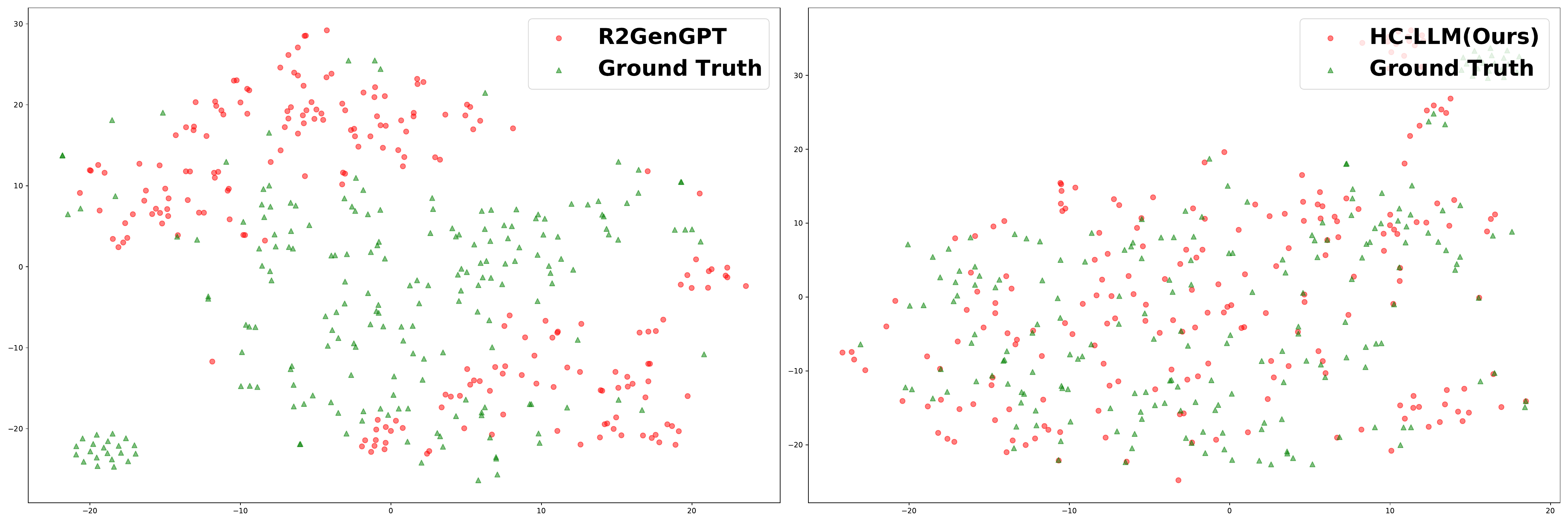}
\vspace{-5mm}
\caption{Visualization of feature distributions using t-SNE for the R2GenGPT and HC-LLM (Ours) on the test set of MS-CXR-T.}
\label{fig 6}
\vspace{-0.3cm}
\end{figure}

As shown in Table~\ref{table 4}, on the MS-CXR-T dataset, our HC-LLM model demonstrates superior performance across various evaluation metrics, further validating the effectiveness and broad applicability of our approach under different architectures. Additionally, we observe that on this smaller dataset, conventional large language models often fail to fully utilize their potential, performing worse in many cases than traditional models designed for single chest x-ray report generation. In contrast, our method, constrained by longitudinal information, shows performance improvements, providing clear evidence of its effectiveness. To further demonstrate the efficacy of our approach compared to other methods on the MS-CXR-T dataset, we visualize the generated reports through t-SNE analysis. As shown in Figure~\ref{fig 6}, the report distributions generated by our method align more closely with the actual report distributions compared to R2GenGPT, providing a clearer and more intuitive validation of its effectiveness.

\subsection{Hyperparameter Analysis}
\begin{figure}[t]
\centering
\includegraphics[width=1.0\linewidth]{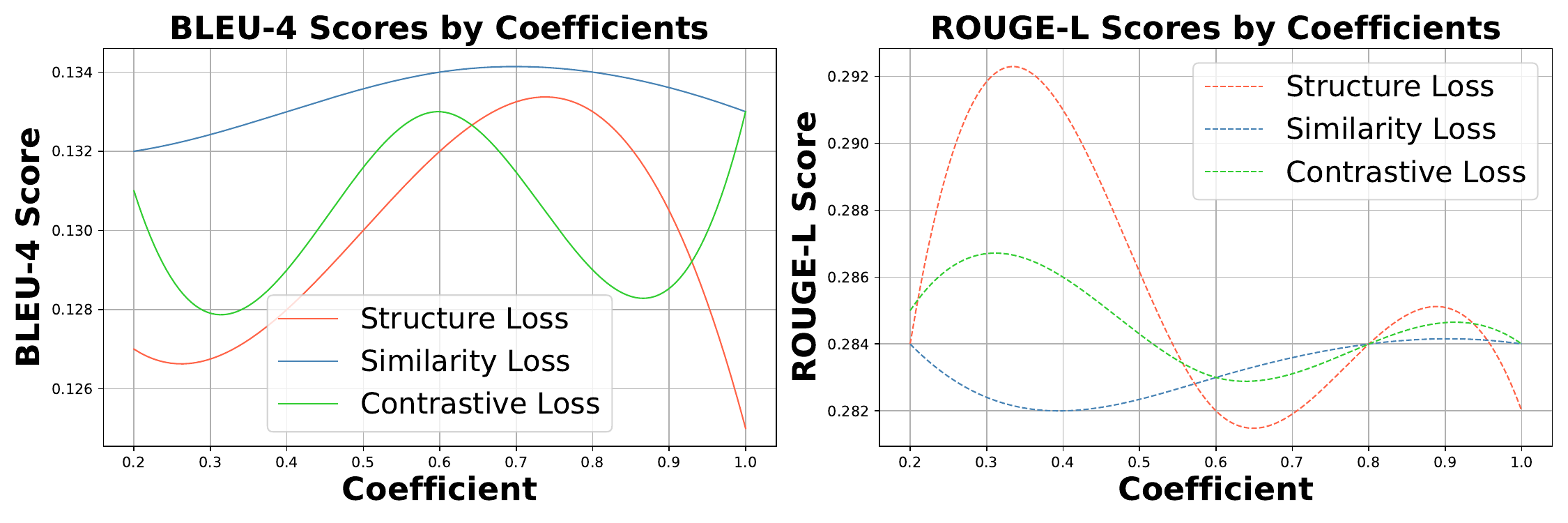}
\vspace{-5mm}
\caption{Performance evaluation of different loss functions across varying coefficients on the MS-CXR-T dataset.}
\label{fig 7}
\vspace{-0.3cm}
\end{figure}

To demonstrate the impact of different constraints on model performance, we conduct experiments on the MS-CXR-T dataset using the R2GenGPT architecture combined with our proposed constraints. The experimental results indicate that the optimal performance of the model is achieved when the structure loss, similarity loss, and contrastive loss are set to 0.8, 0.6, and 1, respectively. As shown in Figure~\ref{fig 7}, the performance curve of the similarity constraint is more stable compared to the structural and contrastive constraints, indicating a smaller impact on overall performance. Our analysis suggests that structural and contrastive constraints can promote the separation of features within and between modalities to a certain extent, reducing reliance on similarity constraints. However, lower similarity may cause the shared and specific features captured by the model to lose their significance, making it difficult for the model to accurately depict changes in disease characteristics, thereby limiting further performance improvements. The fluctuations in contrastive loss performance indicate its crucial role in optimizing feature separation and enhancing report accuracy. At its peak coefficient, contrastive loss effectively distinguishes between features, which is vital for generating precise medical reports. The structural constraint further amplifies this effect by enhancing feature differentiation within and between modalities. Initially, the structural constraint improves performance, but beyond a certain point, they might lead to overfitting, reducing effectiveness. This interplay highlights the need for a carefully balanced approach, ensuring that each constraint complements the others to achieve optimal results on the datasets.

\begin{figure*}[!ht]
\centering
\includegraphics[width=0.9\linewidth]{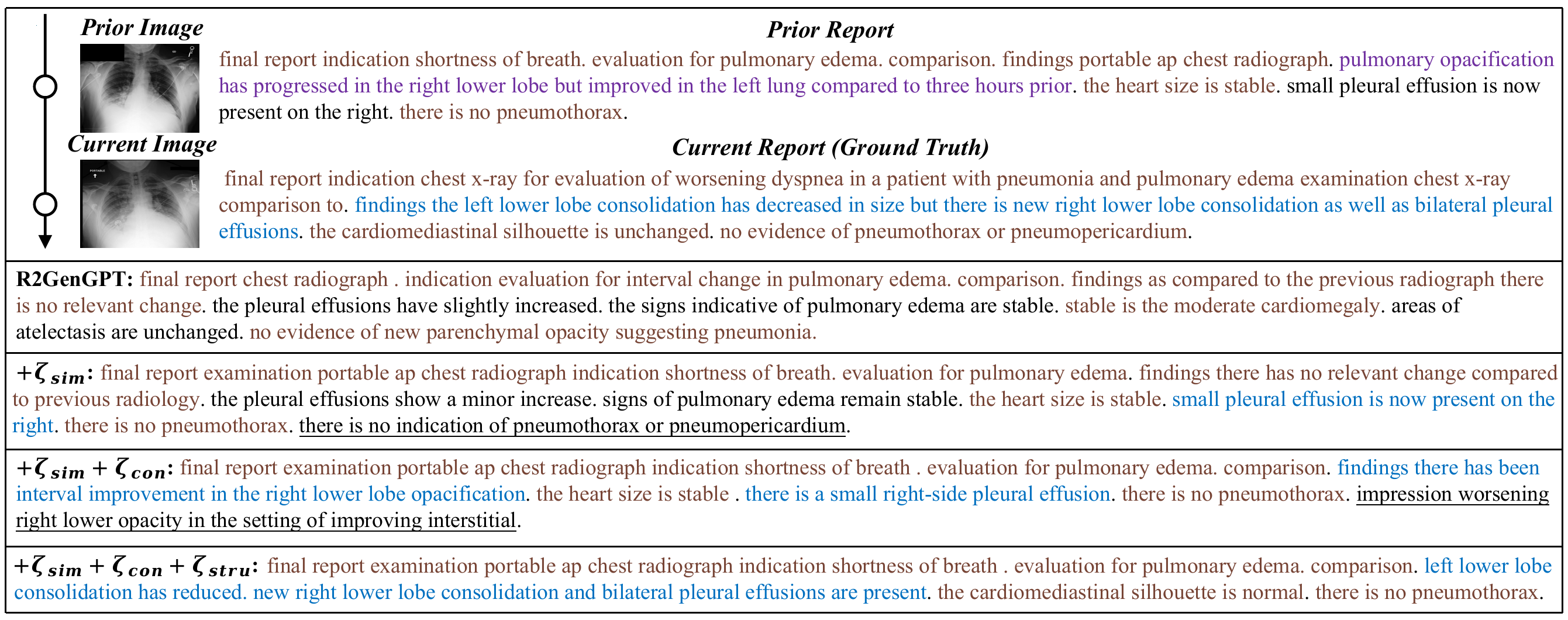}
\caption{An illustration of the reports generated by different models using a sample from the MS-CXR-T test set. Brown represents common content, while purple and blue represent the specific content of each time point. Underlined text indicates incorrect predictions.}
\label{fig 8}
\vspace{-0.3cm}
\end{figure*}

\subsection{More Qualitative Results}
As shown in Figure~\ref{fig 8}, we present the generated reports for a sample from the MS-CXR-T test set. It is evident that there are consistent disease findings across time as well as unique conditions specific to certain times. This underscores our motivation for developing time-shared and time-specific components. For R2GenGPT, the content mainly focuses on common disease information, but contains many inaccuracies. We hypothesize that this is partly due to the limited training data and partly because the model, trained on general data, has lower adaptability to medical datasets, thus relying heavily on historical reports for current report generation. When similarity constraints alone are introduced, there are minimal changes, focusing primarily on generating shared content. With the addition of the contrastive constraint, the model begins to generate both shared and specific report content, benefiting from the strong feature-discriminative power of the contrastive constraint. Finally, by incorporating structural constraint, the results are further optimized, allowing the generated reports to reflect the dynamics of disease progression more accurately, approaching the veracity of actual diagnostic reports.

\subsection{Limitations}
Our limitations can be summarized as follows: 1) The longitudinal radiology report generation task is still in its early stages, with no standard benchmarks available. As a result, we have only validated our results on the Longitudinal-MIMIC and MS-CXR-T datasets. 2) HC-LLM has not yet explored more complex diagnostic data from multiple historical time points, which is crucial for understanding disease progression in clinical diagnoses. In the future, we will conduct further research to advance this task and consider using more realistic multi-historical medical data to further improve its performance.

\end{document}